\documentclass{article}
\usepackage{spconf, amsmath, amsfonts, graphicx}
\usepackage{multirow}
\usepackage{enumitem}
\usepackage{afterpage}
\usepackage[numbers]{natbib}
\usepackage{caption, subcaption}
\usepackage{color, colortbl}
\definecolor{Gray}{gray}{0.9}
\pdfoutput=1

\def \tablescale {0.83}
\def \reducevspace {-0.7em}

\title{Enhancing feature discrimination for Unsupervised Hashing}
\name{Tuan Hoang$^{\star}$ \qquad Thanh-Toan Do$^{\dagger}$ \qquad Dang-Khoa Le Tan$^{\star}$ \qquad Ngai-Man Cheung$^{\star}$}
\address{$^{\star}$Singapore University of Technology and Design (SUTD) \\
    $^{\dagger}$The University of Adelaide}
\begin{document}
%
\maketitle

\begin{abstract}
We introduce a novel approach to improve unsupervised hashing.
Specifically, we propose a very efficient embedding method: \textit{Gaussian Mixture Model embedding (Gemb)}. The proposed method, using Gaussian Mixture Model, embeds feature vector into a low-dimensional vector and, simultaneously, enhances the discriminative property of features before passing them into hashing.
Our experiment shows that the proposed method boosts the hashing performance of many state-of-the-art, e.g. Binary Autoencoder (BA) \cite{BA_7298654}, Iterative Quantization (ITQ) \cite{ITQ_YunchaoGong:2011:IQP:2191740.2191779}, in standard evaluation metrics for the three main benchmark datasets.

\end{abstract}
\vspace{\reducevspace}
\section{Introduction}
\label{sec:intro}

Earlier hashing methods \cite{ITQ_YunchaoGong:2011:IQP:2191740.2191779,BA_7298654,DeepHashing_Liong_2015_CVPR} use   hand-crafted global features, e.g. GIST \cite{GIST_Oliva:2001:MSS:598425.598462}. Recently, the convolutional neural networks (CNNs) have emerged as the state-of-art method for  global descriptors in image retrieval task~\cite{Neural_code,orderless_pooling}. These CNN descriptors inherit the highly discriminative property in visual recognition task. 
Therefore, we hypothesize that: using  the image descriptors based on the activations of CNNs can boost the performance of state-of-the-art hashing methods, compared to hand-crafted features.
Note that  image descriptors from the fully-connected network layers have been evaluated for  hashing \cite{UH-BDNN-Do2016, CBIR_7752525}. However, these works only focus on evaluating the hashing performance of their proposed methods; they do not make any explicit comparison between using hand-crafted and CNN descriptors for hashing.

Given the possibility that the discriminative CNN descriptors can increase the hashing performance of various methods, we delve deeper into the problem: ``\textit{How can we further enhance discrimination of the features for hashing purpose?}'' 
Inspired by state-of-the-art embedding methods including Vector of Locally Aggregated Descriptors \cite{VLAD_JDSP10}, Function Approximation-based Embedding \cite{do_cvpr15,do_tpami2017}, Fisher Vector \cite{FisherVector_4270291}, we propose a method to  enhance the feature discrimination. In particular, different from those embedding methods which transform an incoming variable-size set of independent samples (e.g. SIFT features~\cite{SIFT_Lowe:1999:ORL:850924.851523}) into a higher dimensional fixed size vector representation, our proposed method maps a single descriptor into a lower dimensional fixed-size vector representation. 

\textbf{Contribution.} 
In this paper, 
we address the problem of producing compact but very discriminative features \cite{VLAD_JDSP10}.
Our proposed embedding method improves  hashing performance. 
Specifically, to the best of our knowledge, our work is the first 
to propose to improve unsupervised hashing performance
by an explicit embedding step to 
enhance  feature discrimination.
Our experiments 
show that our embedding method in combination with ITQ \cite{ITQ_YunchaoGong:2011:IQP:2191740.2191779} consistently outperforms other state-of-art hashing methods in  several  evaluation metrics and various datasets.


The remaining of this paper is organized as follows. Section \ref{sec:proposed_method} introduces the proposed method in detail. Section \ref{sec:exp} presents the settings and results of experiments. Finally, we conclude the paper in Section \ref{sec:conclusion}. 
\vspace{\reducevspace}
\section{Proposed method}
\label{sec:proposed_method}

\begin{figure}[t]
\centering
\includegraphics[width=0.48\textwidth]{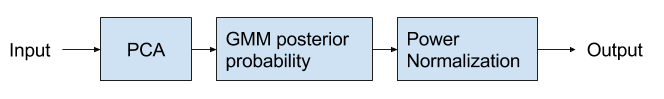}
\label{fig:sparse_chart}
\vspace{-2.0em}
\caption{Our proposed Gemb method: The inputs are global descriptors such as GIST or CNN, and outputs are the embedding features for hashing.}
\end{figure}

In this section, we describe the details of our proposed method with three main steps. In the first step (Section \ref{ssec:proposed_method_dim_reduce}), we pre-process descriptors by Principal Component Analysis (PCA). The main novelty of our method is the second step (Section \ref{ssec:proposed_method_embedding}), in which we attempt to embedding data to lower dimensional space and enhancing feature discrimination simultaneously, using the posterior probabilities of Gaussian Mixture Model (GMM). In the last step (Section \ref{ssec:proposed_method_unsparsifying}), the embedding features are post-processed by Power Normalization to make them more robust to $l2$-distance similarity \cite{PN_Perronnin:2010:IFK:1888089.1888101}. 

\vspace{\reducevspace}
\subsection{Dimensionality reduction}
\label{ssec:proposed_method_dim_reduce}
Our input is a set of $m$ high-dimensional data points:
$\tilde{X}=\{\tilde{x}^{(1)}, \tilde{x}^{(2)}, ..., \tilde{x}^{(m)}\}, \tilde{x}^{(i)} \in \mathbb{R}^d$.
To reduce the computational cost and enhance the discriminative property, we want to produce a compact feature in which the variance of each variable is maximized, the variables are pairwise uncorrelated, and noise and redundancy are removed. This can be accomplished by PCA.  Note that PCA also facilitates our second step, as will be discussed. 

We need to decide the number of PCA components to  retain. 
In this work, we choose the number of retained components based on the percentage of variance retained $\gamma$, $0 \le \gamma \le 1$: 
\vspace{-0.4em}
\begin{equation}
\label{eq:retained_PCA}
\frac{\sum_{j=1}^D \lambda_j}{\sum_{j=1}^n \lambda_j} \ge \gamma.
\end{equation}
Here $\lambda_1, ..., \lambda_n$ are the eigenvalues (sorted in decreasing order) of the covariance matrix $S = \frac{1}{m}\sum_{i=1}^m (\tilde{x}^{(i)})(\tilde{x}^{(i)})^T$.
The number of retained PCA components $D$ is the smallest value that satisfies the inequality (\ref{eq:retained_PCA}). Then, the compressed features can be obtained by: $X = U^T\tilde{X}$, 
where $U$ is the matrix of eigenvectors corresponding to the top-$D$ eigenvalues of $S$ in columns.

Specific to our method, 
reducing the dimension of the data can also reduce the complexity of the hypothesis class considered and help avoid overfitting in learning a GMM (Section \ref{ssec:proposed_method_embedding}). Pairwise uncorrelated variables is also a desirable property to avoid ill-condition covariance matrix in fitting GMM. 
\vspace{-1.0em}
\subsection{Posterior probability as embedding features}
\label{ssec:proposed_method_embedding}
Let $\lambda = \{w_i; \mu_i; \Sigma_i; i = 1...N \}$ be the set of  Gaussian Mixture Model (GMM) parameters learned from the compressed data $X = \{ x^{(1)}, x^{(2)}, ..., x^{(m)}\}; x^{(i)} \in \mathbb{R}^D$. Specifically, $w_i$, $\mu_i$ and $\Sigma_i$ denote respectively the weight, mean vector and covariance matrix of the $i$-th Gaussian in the model
of total $N$ Gaussians.

The posterior probability captures the strength of relationship between a sample $x^{(t)}$ and a Gaussian model $\mathcal{N}(\mu_j, \Sigma_j)$. It is given by:
\begin{equation}
P(j|x^{(t)}, \mu_j, \Sigma_j) = \frac{w_jp_j(x^{(t)}|\mu_j, \Sigma_j)}{\sum\nolimits_{i=1}^Nw_ip_i(x^{(t)}|\mu_i, \Sigma_i)}
\end{equation}
Here 
 $p_j(x^{(t)}|\mu_j, \Sigma_j)$ is the probability of $x^{(t)}$ given the $j$-th Gaussian distribution:
\begin{equation}
p_j(x^{(t)}|\mu_j, \Sigma_j)=\frac{\exp(-\frac{1}{2}(x^{(t)}-\mu_j)^T\Sigma_j^{-1}(x^{(t)}-\mu_j))}{(2\pi)^{D/2}|\Sigma_j|^{1/2}}
\end{equation}
where $|.|$ denotes the determinant operator. 

We propose to construct the embedding feature of a sample $x^{(t)}$ by:
\begin{equation}
z^{(t)}=\left[P(j|x^{(t)}, \mu_j, \Sigma_j); j=1...N \right]
\end{equation}

\begin{figure}[t]
\centering
\begin{subfigure}[b]{0.23\textwidth}
\includegraphics[width=\textwidth]{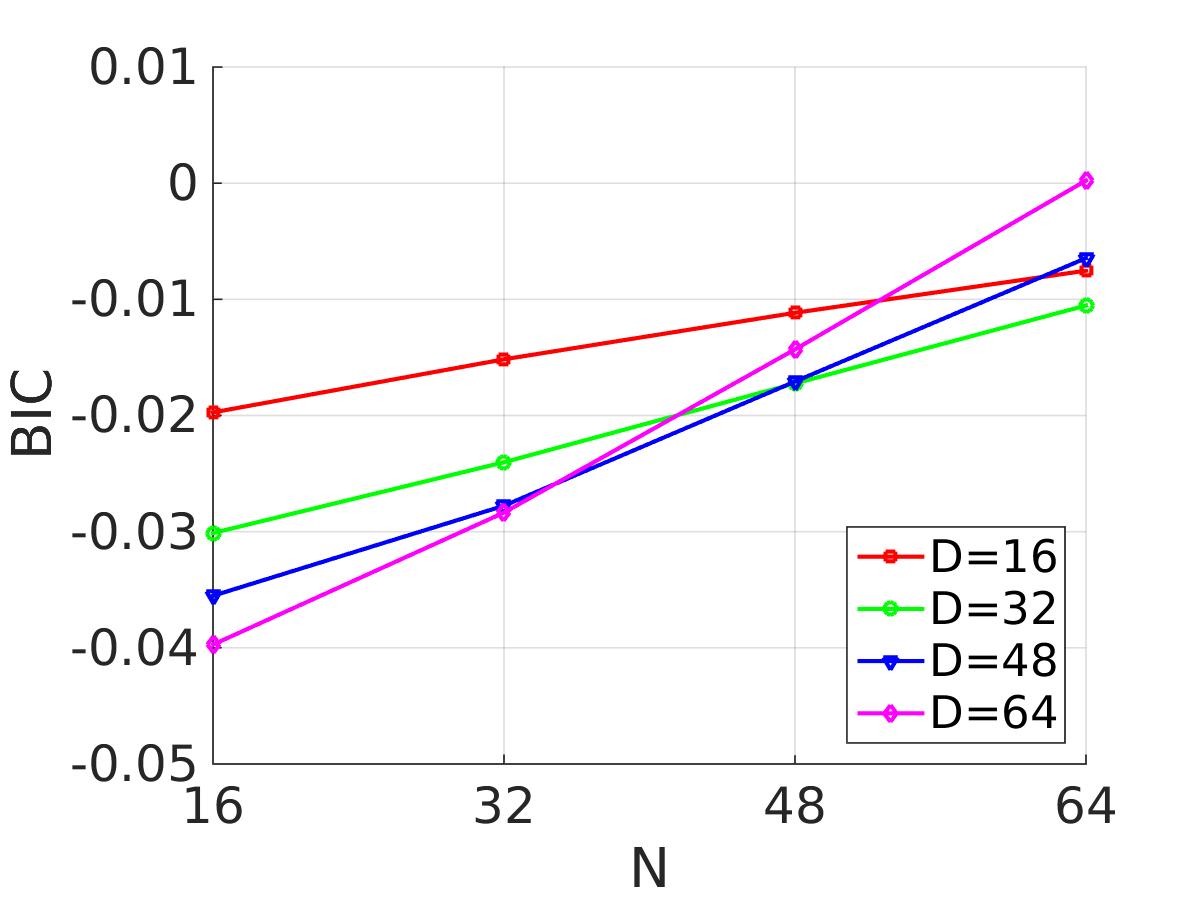}
\caption{CIFAR-10 \cite{cifar10_krizhevsky2009learning}}
\end{subfigure}
\begin{subfigure}[b]{0.23\textwidth}
\includegraphics[width=\textwidth]{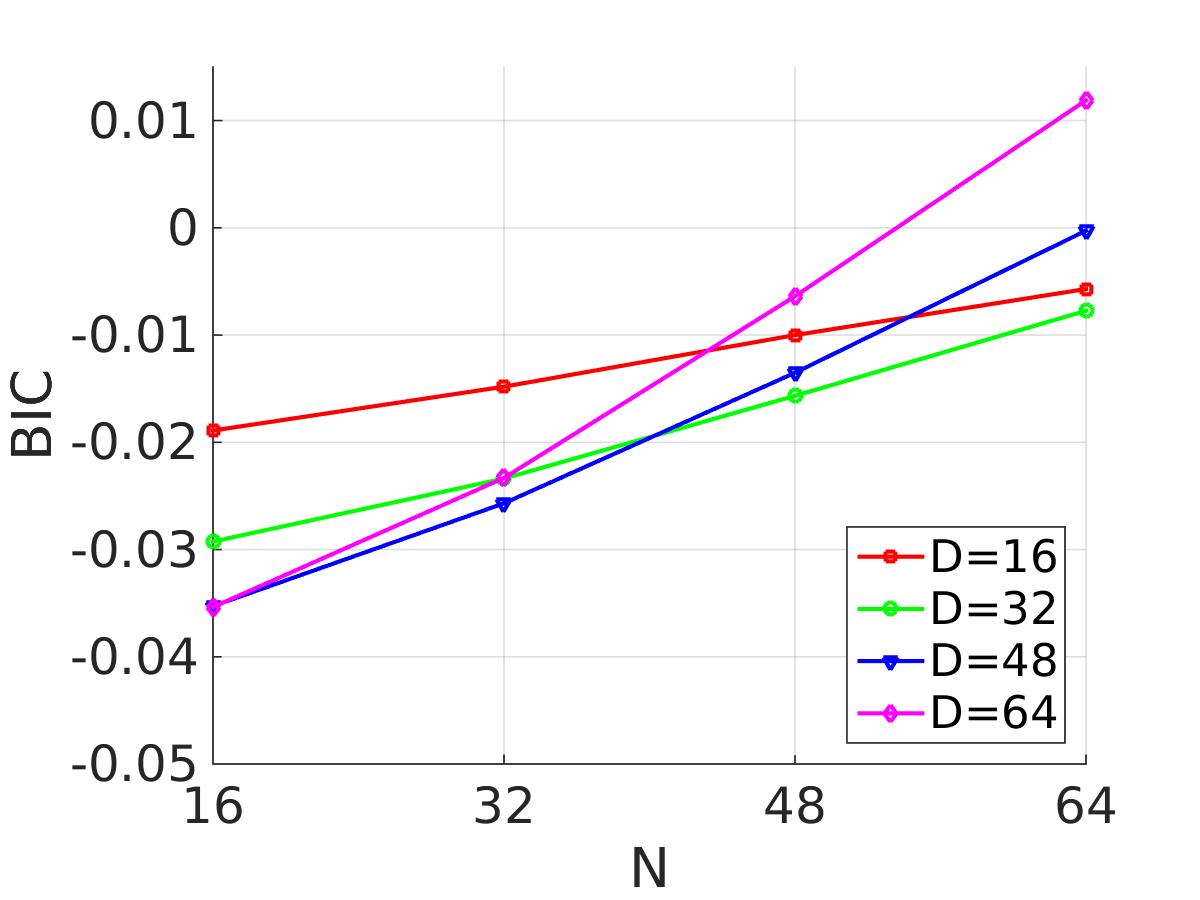}
\caption{LabelMe-12-50k \cite{labelme_Uetz_large-scaleobject}}
\end{subfigure}
\caption[]{The relative difference between BIC values of GMM with full covariances and GMM with diagonal covariances using VGG-FC7 descriptors of CIFAR-10 and LabelMe-12-50k dataset\footnote{Please refer to Section 3.1 for more information about the datasets.}. A negative difference means that the former one is better under BIC.}
\label{fig:BIC_gap}
\end{figure}

In our proposed method, since we do not have the diagonal constraint on the covariances as in Fisher Vector encoding \cite{FisherVector_4270291}, we decide the use full covariances in the GMM to achieve better mixture models. With full covariances instead of diagonal covariances, it is possible to increase the likelihood.  However, there is a risk of overfitting as there are more parameters to be estimated. 
Therefore, in order to make a systematic comparison between GMM with full and diagonal covariances, we utilize the Bayesian Information Criterion (BIC) \cite{BIC_STAN:STAN530} as this criterion reduces the risk of overfitting by introducing the penalty term on the number of parameters.
BIC results are shown in Fig.\ref{fig:BIC_gap}.  We observe that with a small $N$, e.g. 16, using full covariances can help  achieve much better mixture models under BIC (or smaller BIC). At a larger $N$, e.g. 64, the BIC values of GMM with diagonal covariances are more comparable to, or even higher than (at large $D$) those of GMM with full covariances. Therefore, full covariance is preferable.

\footnotetext{Please refer to Section \ref{ssec:dataset_evalProtocol} for detail information about the datasets.}
\vspace{\reducevspace}
\subsection{Unsparsifying by Power Normalization}
\label{ssec:proposed_method_unsparsifying}

Similar to the finding in \cite{PN_Perronnin:2010:IFK:1888089.1888101}, we observe that as the number of Gaussians increases, the embedding features become sparser. The distributions of  embedding features (log-scale) in Fig. \ref{fig:sparse_chart} shift to more negative regions as $N$ increases.  This effect can be explained: as the number of Gaussians increases, it is easier for a embedding feature $z^{(t)}$ to fit in a distribution with high probability.


Additionally, we also observe in Fig. \ref{fig:sparse_chart} that  more discriminative   global descriptors $\tilde{x}^{(t)}$ result in  sparser  embedding features $z^{(t)}$. This property can be explained: with more discriminative descriptors,  the descriptors of semantically similar images tend to be very close in the $l2$-space, while those of semantically different images tend to locate far away from each other. Consequently, GMM groups the semantically similar descriptors into the same clusters with high probability.


\begin{figure}[h]
\centering

\begin{subfigure}[b]{0.23\textwidth}
\includegraphics[width=\textwidth]{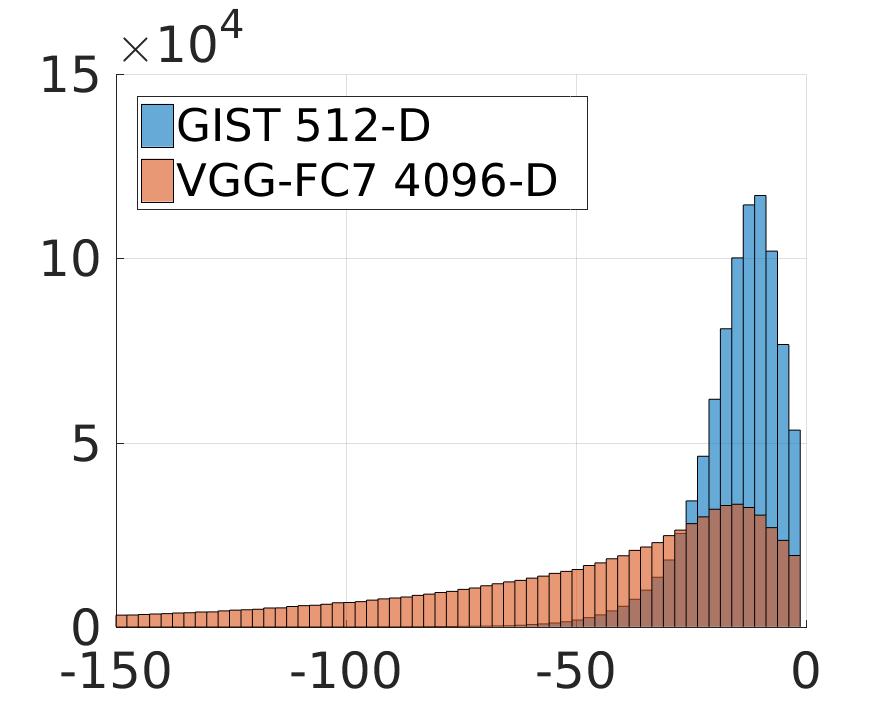}
\caption{$N=16$}
\end{subfigure}
\begin{subfigure}[b]{0.23\textwidth}
\includegraphics[width=\textwidth]{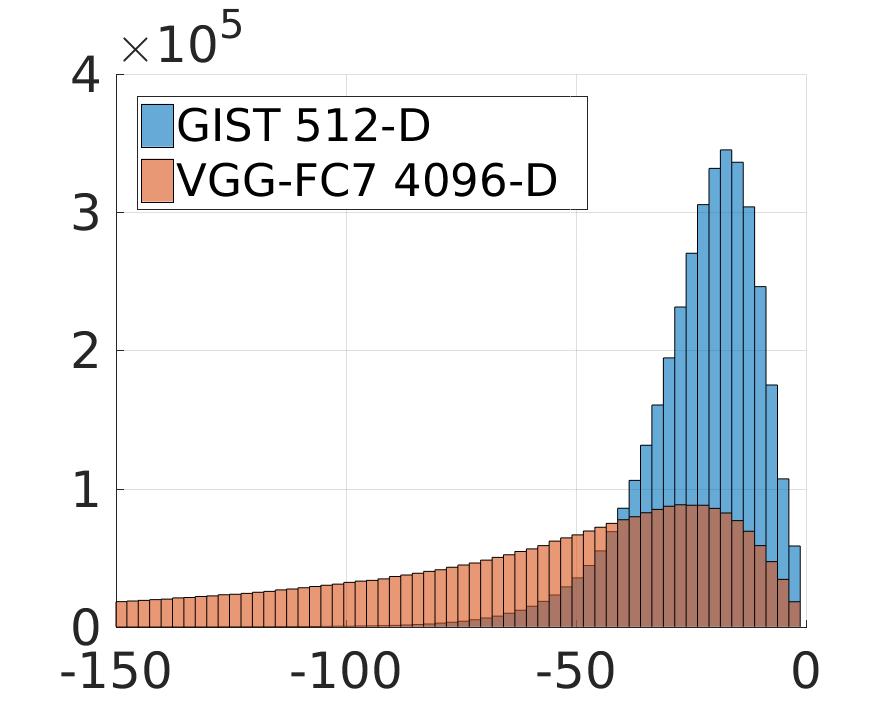}
\caption{$N=64$}
\end{subfigure}
\caption{Histogram of embedding features (in log scale) for CIFAR-10  \cite{cifar10_krizhevsky2009learning}  using GIST 512-D and VGG-FC7 4096-D descriptors$^1$. Horizontal-axis is in log-scale.}
\label{fig:sparse_chart}
\end{figure}

With sparse vectors, the $l2$-distance is a poor measurement of similarity.  Therefore, we follow \cite{PN_Perronnin:2010:IFK:1888089.1888101} to ``\textit{unsparsify}'' the embedding features by applying the Power Normalization function (\ref{eq:power_norm}), so that we can still use the $l2$-distance similarity. In particular, the $l2$-distance similarity is desirable as  hashing methods preserve the similarity between Euclidean and Hamming distances. 
\begin{equation}
\label{eq:power_norm}
f(z)=\text{sign}(z)|z|^\alpha
\end{equation}

\vspace{-1.0em}
\vspace{\reducevspace}
\subsection{Visualize descriptors}
\label{ssec:visualize_desc}

\begin{figure}[h]
\centering

\begin{subfigure}[b]{0.156\textwidth}
\includegraphics[width=\textwidth]{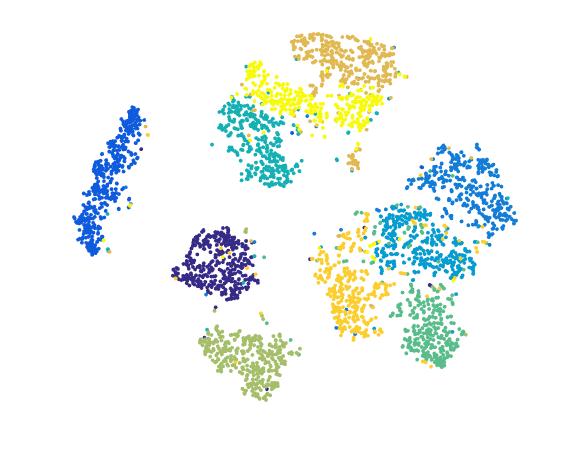}
\caption{Without Gemb}
\end{subfigure}
\begin{subfigure}[b]{0.156\textwidth}
\includegraphics[width=\textwidth]{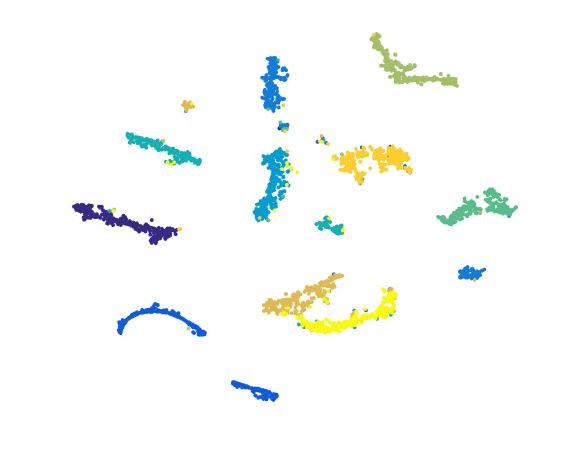}
\caption{Gemb ($N=32$)}
\label{fig:gemb_32}
\end{subfigure}
\begin{subfigure}[b]{0.156\textwidth}
\includegraphics[width=\textwidth]{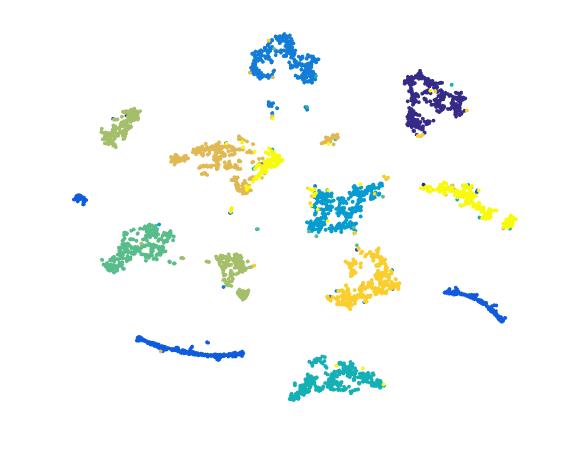}
\caption{Gemb ($N=64$)}
\label{fig:gemb_64}
\end{subfigure}
\caption{Visualizing GIST descriptors of a subset of MNIST dataset$^1$. (a): without Gemb.  (b, c): with Gemb.  Different colors correspond to different classes. Best viewed in color.}
\label{fig:mnist_cluster}
\end{figure}

We utilize t-SNE \cite{t-sne_van2008visualizing} to visualize (Fig. \ref{fig:mnist_cluster}) the scatter plot of GIST 512-D descriptors of a subset of MNIST dataset. We can clearly observe that after processing with Gemb with a certain number of Gaussians (Fig. \ref{fig:gemb_32}, \ref{fig:gemb_64}), the descriptors of the same class locate closer.
Furthermore, the boundaries between different classes are clearer. These suggest that Gemb can lead to  more discriminative features.

\vspace{\reducevspace}
\vspace{-1.0em}
\section{Experiments}
\label{sec:exp}

Firstly, we conduct comprehensive experiments to show that, in comparison with GIST 512-D descriptors \cite{GIST_Oliva:2001:MSS:598425.598462}, descriptors from CNN (e.g. VGG-FC7 \cite{VGG_DBLP:journals/corr/SimonyanZ14a}) can significantly boost the retrieval performance of many state-of-the-art unsupervised hashing methods: Spectral Hashing (SH) \cite{SH_NIPS2008_3383}, Spherical Hashing (SpH) \cite{SpH_6248024}, Iterative Quantization (ITQ) \cite{ITQ_YunchaoGong:2011:IQP:2191740.2191779}, and Binary Autoencoder (BA) \cite{BA_7298654} (Section \ref{ssec:exp_gist_vs_cnn}). More importantly, we demonstrate that our proposed method in combination with ITQ and BA further enhance the hashing quality for both GIST hand-crafted features and CNN descriptors (Section \ref{ssec:evaluate_gemb}).

\vspace{\reducevspace}
\subsection{Dataset, Evaluation protocol, and Implementation notes}
\label{ssec:dataset_evalProtocol}
The \textbf{CIFAR-10 dataset}~\cite{cifar10_krizhevsky2009learning} contains 60,000  fully-annotated color images of $32 \times 32$ from 10 object classes. We randomly sampled 10\% images per class, as the query data, and used the remaining images as the training set and retrieval database. This sampling strategy is applied to all other datasets to handle unbalanced samples in different classes.

The \textbf{LabelMe-12-50k dataset} \cite{labelme_Uetz_large-scaleobject} is a subset of LabelMe dataset \cite{labelme_full_Russell:2008:LDW:1345995.1345999}. The LabelMe-12-50k dataset includes 50,000 fully-annotated color images of $256 \times 256$ of 12 classes. In this dataset, for images which have more than one label values in $[0.0, 1.0]$, we choose the object class corresponding to the largest label value as image labels.


The \textbf{MNIST dataset}~\cite{lecun-mnisthandwrittendigit-2010} consists of 70,000 fully-annotated grayscale handwritten digit images of $28\times 28$ from 10 classes.

\textbf{Evaluation protocols}. To evaluate retrieval performance of methods, we apply three common metrics: 
1) mean Average Precision (\textbf{\textit{mAP}}); 
2) precision of Hamming radius of 2 (\textbf{\textit{precision@r2}}) which measures precision on retrieved images having Hamming distance to query $\le 2$ (we report zero precision for the queries that return no image); 
3) precision at top 1000 return images (\textbf{\textit{precision@1k}}) which measures the precision on the top 1000 retrieved images. 
Class labels are used as ground truths for all evaluation.
Additionally, to avoid biased results due to unbalanced samples of different classes in query sets, we calculate the average results for all classes. 
\smallskip

\textbf{Implementation notes.} In our method, the two parameters $\{\gamma, \alpha\}$ is empirically set as $\{0.85, 0.15\}$ and $\{0.65, 0.05\}$ for GIST 512-D \cite{GIST_Oliva:2001:MSS:598425.598462} and VGG-FC7
\footnote{Descriptors are extracted from the Fully Connected layer 7 of VGG.}
\cite{VGG_DBLP:journals/corr/SimonyanZ14a} descriptors respectively. The number of Gaussians $N$ is set as the number of hashing bits.

For all compared methods \cite{BA_7298654,ITQ_YunchaoGong:2011:IQP:2191740.2191779,SH_NIPS2008_3383,SpH_6248024}, we use the implementations and the suggested parameters provided by the authors. Besides, in order to improve the statistical stability in the results, we execute the experiments 5 times
for all methods and report the average values. 
\vspace{\reducevspace}
\subsection{Hand-crafted descriptor vs. CNN descriptor}
\label{ssec:exp_gist_vs_cnn}

Table \ref{tb:cifar10} and Table \ref{tb:labelme} compare the hashing performance of different methods \cite{SH_NIPS2008_3383,SpH_6248024, ITQ_YunchaoGong:2011:IQP:2191740.2191779,BA_7298654} using GIST 512-D descriptors and the activation of the fully connected layer of VGG (VGG-FC7). As shown in the tables, all hashing methods consistently achieve higher performances in the majority of evaluation metrics when using the VGG-FC7 descriptors in comparison with GIST 512-D descriptors. 

\begin{table}[t]

\caption{Results on the CIFAR-10 \cite{cifar10_krizhevsky2009learning} dataset. We report the results using GIST 512-D descriptor \cite{GIST_Oliva:2001:MSS:598425.598462} on the top section and using VGG-FC7 descriptor \cite{VGG_DBLP:journals/corr/SimonyanZ14a} on the bottom section.}
\label{tb:cifar10}
\centering
\scalebox{\tablescale}{
\begin{tabular}{|l|c|c|c|c|c|c|c|}
\hline
\multirow{2}{*}{Methods} 
& \multicolumn{3}{c|}{mAP} & \multicolumn{2}{c|}{precision@1k} & \multicolumn{2}{c|}{precision@r2}  \\ \cline{2-8}
& 16 & 32 & 64 & 16 & 32 & 16 & 32 \\ 
\hline


SH \cite{SH_NIPS2008_3383} &12.88 & 12.71 & 12.99 & 17.32 & 17.69 & 18.38 & 21.00 \\

SpH \cite{SpH_6248024} & 14.46 & 15.13 & 15.88 & 19.38 & 21.30 & 21.20 & 13.86\\
\rowcolor{Gray}
BA \cite{BA_7298654} & 15.34 & 16.86 & 17.74 & 21.64 & 24.30 & 24.65 & 13.67\\

ITQ \cite{ITQ_YunchaoGong:2011:IQP:2191740.2191779} & 16.59 & 17.42 & 18.02 & 22.36 & 24.49 & 23.84 & 18.24 \\
\rowcolor{Gray}
Gemb+BA & 20.80 & 22.20 & 22.45 & 28.86 & \textbf{32.86} & \textbf{28.31} & 32.58 \\

Gemb+ITQ & \textbf{21.36} & \textbf{22.44} & \textbf{22.59} & \textbf{29.02} & 31.48 & 28.25 & \textbf{32.97} \\
\hline
SH \cite{SH_NIPS2008_3383} & 18.31 & 16.54 & 15.78 & 28.61 & 26.74 & 32.90 & 18.95 \\

SpH \cite{SpH_6248024} & 18.82 & 20.93 & 23.40 & 27.33 & 31.60 & 30.34 & 22.33 \\
\rowcolor{Gray}
BA \cite{BA_7298654} & 25.38 & 26.16 & 27.99 & 36.45 & 38.19 & \textbf{39.58} & 25.54 \\

ITQ \cite{ITQ_YunchaoGong:2011:IQP:2191740.2191779} & 26.82 & 27.38 & 28.73 & \textbf{37.08} & 38.86 & 38.83 & 29.53 \\
\rowcolor{Gray}
Gemb+BA & 27.24 & 28.52 & 29.97 & 36.37 & \textbf{39.64} & 35.45 & \textbf{42.05} \\

Gemb+ITQ & \textbf{27.61} & \textbf{29.12} & \textbf{30.01} & 35.51 & 38.36 & 34.03 & 41.26 \\
\hline
\end{tabular}
}
\end{table}

\begin{table}[ht]

\caption{Results on the LabelMe-12-50k dataset \cite{labelme_Uetz_large-scaleobject}. We report the results using GIST 512-D descriptor \cite{GIST_Oliva:2001:MSS:598425.598462} on the top section and using VGG-FC7 descriptor \cite{VGG_DBLP:journals/corr/SimonyanZ14a} on the bottom section.}
\label{tb:labelme}
\centering
\scalebox{\tablescale}{
\begin{tabular}{|l|c|c|c|c|c|c|c|}
\hline
\multirow{2}{*}{Methods} 
& \multicolumn{3}{c|}{mAP} & \multicolumn{2}{c|}{precision@1k} & \multicolumn{2}{c|}{precision@r2}  \\ \cline{2-8}
& 16 & 32 & 64 & 16 & 32 & 16 & 32 \\ 
\hline


SH \cite{SH_NIPS2008_3383} & 10.74 & 10.76 & 10.95 & 13.73 & 13.82 & 14.52 & 18.54 \\

SpH \cite{SpH_6248024} & 11.86 & 13.02 & 13.67 & 15.17 & 17.07 & 16.96 & 13.89\\

\rowcolor{Gray}
BA \cite{BA_7298654} & 14.21 & 14.55 & 15.43 & 18.93 & 19.97 & 19.92 & 12.97\\

ITQ \cite{ITQ_YunchaoGong:2011:IQP:2191740.2191779} & 15.07 & 16.06 & 16.58 & 18.83 & 20.31 & 20.24 & 19.43 \\
\rowcolor{Gray}
Gemb+BA & 19.95 & 21.64 & 21.69 & 24.43 & \textbf{26.33} & 24.56 & 27.92 \\

Gemb+ITQ & \textbf{20.79} & \textbf{21.69} & \textbf{21.92} & \textbf{24.77} & 26.01 & \textbf{24.75} & \textbf{28.67} \\
\hline
SH \cite{SH_NIPS2008_3383} & 12.60 & 12.59 & 12.24 & 17.23 & 17.20 & 20.32 & 15.23 \\

SpH \cite{SpH_6248024} & 13.59 & 15.10 & 17.03 & 17.81 & 20.08 & 20.39 & 14.29 \\

\rowcolor{Gray}
BA \cite{BA_7298654} & 16.96 & 18.42 & 20.80 & 21.99 & 23.85 & 25.80 & 15.40 \\

ITQ \cite{ITQ_YunchaoGong:2011:IQP:2191740.2191779} & 18.06 & 19.40 & 20.71 & 23.13 & 24.84 & \textbf{26.30} & 19.54 \\
\rowcolor{Gray}
Gemb+BA & 22.63 & 24.05 & 24.19 & \textbf{27.15} & 28.47 & 25.95 & \textbf{30.77} \\

Gemb+ITQ & \textbf{23.37} & \textbf{24.26} & \textbf{25.37} & 25.65 & \textbf{28.92} & 24.91 & 29.38 \\
\hline
\end{tabular}
}
\end{table}

\begin{table}[ht]

\caption[]{Results on the MNIST \cite{lecun-mnisthandwrittendigit-2010} dataset. We report the results using GIST 512-D descriptors \cite{GIST_Oliva:2001:MSS:598425.598462}.}
\label{tb:mnist}
\centering
\scalebox{\tablescale}{
\begin{tabular}{|l|c|c|c|c|c|c|c|}
\hline
\multirow{2}{*}{Methods} 
& \multicolumn{3}{c|}{mAP} & \multicolumn{2}{c|}{precision@1k} & \multicolumn{2}{c|}{precision@r2}  \\ \cline{2-8}
& 16 & 32 & 64 & 16 & 32 & 16 & 32 \\ 
\hline


SH \cite{SH_NIPS2008_3383} & 32.59 & 33.23 & 30.65 & 56.23 & 61.03 & 60.12 & 78.37 \\

SpH \cite{SpH_6248024} & 31.27 & 36.80 & 41.40 & 51.28 & 62.17 & 57.66 & 68.62\\
\rowcolor{Gray}
BA \cite{BA_7298654} & 48.48 & 51.72 & 52.73 & 70.83 & 76.23 & 75.17 & 74.70\\

ITQ \cite{ITQ_YunchaoGong:2011:IQP:2191740.2191779} & 46.37 & 50.59 & 53.69 & 69.29 & 75.85 & 70.66 & 82.06 \\
\rowcolor{Gray}
Gemb+BA & 79.35 & 82.59 & 83.13 & 85.66 & 92.74 & 84.56 & 92.56 \\

Gemb+ITQ & \textbf{79.97} & \textbf{83.81} & \textbf{83.72} & \textbf{86.09} & \textbf{93.33} & \textbf{85.13} & \textbf{93.23} \\



\hline
\end{tabular}
}

\end{table}
\vspace{\reducevspace}
\subsection{Evaluate Gemb}
\label{ssec:evaluate_gemb}

In order to evaluate Gemb for hashing purpose, we combine Gemb with BA and ITQ\footnote{Due to space constraint, we only evaluate our Gemb with BA and ITQ. However, Gemb also works well with other hashing methods \cite{SH_NIPS2008_3383, SpH_6248024, KMH,UH-BDNN-Do2016}.} and conduct experiments on CIFAR-10, LabelMe-12-50k, and MNIST\footnote{We do not evaluate the descriptor from VGG for MNIST dataset since VGG is trained on RGB images while MNIST is grayscale.} datasets. We report experimental results for these datasets on Table \ref{tb:cifar10}, Table \ref{tb:labelme}, and Table \ref{tb:mnist} respectively. Gemb clearly helps to boost performance of BA and ITQ in majority of evaluation metrics, \textbf{\textit{mAP}}, \textbf{\textit{precision@1k}} and \textbf{\textit{precision@r2}}, for various code lengths and datasets. Furthermore, Gemb+ITQ consistently achieves the best \textbf{\textit{mAP}} among all compared methods. 

\smallskip
\textbf{Comparison with Hashing method using Deep Neural Network (DNN).} Recently, there are several methods \cite{DeepHashing_Liong_2015_CVPR,UH-BDNN-Do2016,DeepBit_LinLearningCB} to apply DNN to learn binary hash code. These methods achieve very competitive performances. 

\begin{itemize}[leftmargin=*]
\item \textbf{\textit{Deep Hasing (DH) \cite{DeepHashing_Liong_2015_CVPR}}} and \textbf{\textit{Unsupervised Hashing with Binary Deep Neural Network (UH-BDNN) \cite{UH-BDNN-Do2016}.}}
Following the experiments settings in \cite{DeepHashing_Liong_2015_CVPR} and \cite{UH-BDNN-Do2016}, we conduct experiments on CIFAR-10 to make a fair comparison. In this experiment, 100 images are randomly sampled for each class as query set; the remaining images are for training and database query. The images are presented by GIST 512-D descriptors. 
In addition, to avoid bias results due to test samples, we repeat the experiment 5 times with 5 different random test sets. 
The comparative results in term of \textbf{\textit{mAP}} and \textbf{\textit{precision@r2}} are presented in Table \ref{tb:DH-UH_BDNN}. Clearly, Gemb + ITQ consistently outperforms DH and UH-BDNN.

\begin{table}[ht]
\caption{Comparison with Deep Hashing (DH) \cite{DeepHashing_Liong_2015_CVPR} and Unsupervised Hashing with Binary Deep Neural Network (UH-BDNN) \cite{UH-BDNN-Do2016}. The results of DH and UH-BDNN are cited from \cite{DeepHashing_Liong_2015_CVPR} and \cite{UH-BDNN-Do2016} respectively.}
\label{tb:DH-UH_BDNN}
\centering
\scalebox{\tablescale}{
\begin{tabular}{|l | c| c| c| c|}
\hline
\multirow{3}{*}{Methods} & \multicolumn{4}{c|}{CIFAR10} \\ \cline{2-5}
& \multicolumn{2}{c|}{mAP} & \multicolumn{2}{c|}{precision@r2}  \\ \cline{2-5}
& 16 & 32 & 16 & 32 \\ 
\hline
DH~\cite{DeepHashing_Liong_2015_CVPR} & 16.17 & 16.62 &23.33 &15.77 \\

UH-BDNN~\cite{UH-BDNN-Do2016} & 17.83 &18.52& 24.97& 18.85 \\

Gemb+ITQ & \textbf{21.14} & \textbf{22.27} & \textbf{28.60} & \textbf{31.18} \\
\hline


\end{tabular}
}
\end{table}
\item \textbf{\textit{DeepBit~\cite{DeepBit_LinLearningCB}.}}
Similarly, we compare Gemb + ITQ with DeepBit on CIFAR-10. In this experiment, since DeepBit utilized pretrained VGG to learn the compact binary code, we present each image by a VGG-FC7 descriptor. We report the \textbf{\textit{mAP}} of the top 1,000 returned images in Table \ref{tb:deepbit} as in \cite{DeepBit_LinLearningCB}. The results clearly show that Gemb + ITQ outperforms DeepBit by a fair margin.

\begin{table}[ht]
\caption{Comparison with DeepBit \cite{DeepBit_LinLearningCB}. The results of DeepBit  are cited from \cite{DeepBit_LinLearningCB}. }
\label{tb:deepbit}
\centering
\scalebox{\tablescale}{
\begin{tabular}{|l | c| c| c| }
\hline
Methods & 16 & 32 & 64 \\
\hline
DeepBit \cite{DeepBit_LinLearningCB} & 19.43 & 24.86 & 27.73  \\ 
Gemb+ITQ& \textbf{37.50} & \textbf{41.00} & \textbf{44.27} \\
\hline
\end{tabular}
}
\end{table}
\end{itemize}

\vspace{\reducevspace}
\section{Conclusion}
\label{sec:conclusion}
We have discussed a novel approach to improve unsupervised hashing by an explicit embedding step to enhance feature discrimination.
In particular, we have discussed our proposed Gemb method using GMM posterior probabilities.  Our method attempts to embed a high-dimensional global descriptor into a low-dimensional space and, simultaneously, enhance the embedding feature discrimination. The solid experimental results on three benchmark datasets have demonstrated that Gemb can help to enhance feature discrimination and boost the performances of different hashing methods. 

\clearpage
\bibliographystyle{IEEEbib}
\bibliography{refs}

\end{document}